\documentclass{article}

\usepackage{microtype}
\usepackage{graphicx}
\usepackage{subfigure}
\usepackage{booktabs}
\usepackage{amsmath,amssymb}

\usepackage{hyperref}

\usepackage[accepted]{icml2019}

\icmltitlerunning{World Programs for Model-Based Learning and Planning}

\begin{document}

\twocolumn[
\icmltitle{World Programs for Model-Based Learning and Planning in Compositional State and Action Spaces}




\begin{icmlauthorlist}
\icmlauthor{Marwin H.S. Segler}{add,ms}
\end{icmlauthorlist}

\icmlaffiliation{add}{BenevolentAI, London, UK}
\icmlaffiliation{ms}{Westf\"alische Wilhelms-Universit\"at M\"unster, DE}

\icmlcorrespondingauthor{}{marwin.segler@wwu.de}

\icmlkeywords{Machine Learning, ICML}

\vskip 0.3in
]



\printAffiliationsAndNotice{}  

\begin{abstract}

Some of the most important tasks take place in environments which lack cheap and perfect simulators, thus hampering the application of model-free reinforcement learning (RL). 
While model-based RL aims to learn a dynamics model, in a more general case the learner does not know a priori what the action space is.
Here we propose a formalism where the learner induces a world program by learning a dynamics model and the actions in graph-based compositional environments by observing state-state transition examples. Then, the learner can perform RL with the world program as the simulator for complex planning tasks. 
We highlight a recent application, and propose a challenge for the community to assess world program-based planning.
\end{abstract}

\section{Introduction}
Consider environments represented by a Markov Decision Process $\mathcal{M} = (\mathcal{S},\mathcal{A}, \mathcal{T},\mathcal{R})$ with discrete states $\mathcal{S}$ and actions $\mathcal{A}$. 
Deep Reinforcement Learning has been successful in tasks with moderate state and action space sizes where cheap and perfect environment simulators $\mathcal{M}^\prime$ are available, or where the agent can even be placed in the real environment $\mathcal{M}$, as in many games. Agents can then be trained by interacting many times with the (simulated) environment.

Unfortunately, in many important tasks of real life interest, this framework cannot be easily applied.
First, interactions with the real environment can be very expensive, and even potentially dangerous. Obviously, it would be prohibitive to let an agent initially drive randomly on public roads, or to perform unrestricted scientific wet-lab experiments, which can cost 1000s of pounds \cite{rwrl}.

It has been recognised early that it is possible to train models as simulators of the environment dynamics $\mathcal{T^\prime}$ based on limited interaction data, for example recorded trajectories or experiments which have been performed in the past \cite{juergen1,ha2018recurrent}. However, canonical model-based RL still requires that the action space $\mathcal{A}$ is known and can be provided easily by the programmer.

More generally, the learner does not have information available about the possible actions they can take, since it can be too complex to specify what $\mathcal{A}$ actually is, similar to common sense knowledge in expert systems. The learner then first has to learn the nature of the elements of the set $\mathcal{A}$ such that they are compliant with observations $(s_t, s_{t+1})$ of subsequent states. 

Second, in contrast to board games or image representations which are of fixed size, in a more general setting the state space $\mathcal{S}$ is compositional, constructed of multiple entities and their relations, which means it can potentially have countably infinite cardinality $|\mathcal{S}| \leq \infty$, which requires the learner to exhibit an appropriate inductive bias. Agents with representations based on graphs can naturally tackle compositionality in environments, and have been successful in chemical \cite{segler2017towards,segler2018planning} and physical \cite{hamrick2018relational,bapst2019structured} reasoning and planning tasks.

In the frontiers section of their book \citet{RLbook} succinctly state that 
\textit{``we still need scalable methods for planning with learned environment models. 
Planning methods have proven extremely effective in applications such as AlphaGo Zero and computer chess in which the model of the environment is known from the rules of the game or can otherwise be supplied by human designers. 
But cases of full model-based reinforcement learning, in which the environment model is learned from data and then used for planning, are rare.''}

Here, we describe a general formalism to address this issue for learning and planning in compositional state spaces with a priori unknown actions, as merger of three components: 1) The induction of the action space as a graph-based world program from a small set of state-state transitions, 2) learning of neural network policies and dynamics models, and 3) planning using these networks and the world program. We further highlight its successful application, and future challenges for the RL community.

\section{Formalism}
Consider the problem of near-optimal decision making in infinite, discrete, compositional state and action spaces, where a simulator of the environment dynamics $\mathcal{T}$ is not available. Furthermore, consider that the learner does not know the elements of the action space $\mathcal{A}$ a priori. The only clue are limited amounts of batch data in the form of state-state pairs $(s_t, s_{t+1})$ from (previous) observations of the environment. In the framework of Markov Decision Processes \cite{RLbook} we then have
\begin{align}
&\overset{?}{\mathcal{A}} &\text{\tiny actions}\\
\mathcal{A}(s) &: \mathcal{S} \overset{?}{\rightarrow} \wp(\mathcal{A}) &\text{\tiny available actions in state $s$}\\
\pi &: \mathcal{S} \overset{?}{\rightarrow} \mathcal{A} &\text{\tiny policy}\\
\mathcal{T} &:  { } \mathcal{S} \times \mathcal{A} \times \mathcal{S} \overset{?}{\rightarrow} [0,1] &\text{\tiny dynamics}\\
\mathcal{R} &: { } \mathcal{S} \times \mathcal{A} \rightarrow \mathbb{R} &\text{\tiny reward}
\end{align}
where $\overset{?}{\mathcal{A}}$ or $\overset{?}{\rightarrow}$ indicate that this set or mapping has to be learned. For the sake of simplicity, we assume $\mathcal{R}$ is given.\footnote{One could consider learning the reward function via inverse RL \cite{ng2000algorithms}.} 

\subsection{Graph-Based Environments}
We consider compositional environments, where states are composed of multisets of graphs.\footnote{With some loss of generality, these sets of graphs may be merged into a single graph with multiple disconnected components. However, this makes it more challenging to treat different groups of disconnected components.} 
Graphs $G=(V,E)$ in turn are composed of vertices (nodes) $V$  and edges $E$, which model the relationship between the vertices. 
Graphs may be vertex- or edge-labeled with arbitrarily structured labels, and can be directional. 
This subsumes for example grids (lattice graphs, images), text (linear chain graphs), computer code (abstract syntax trees), molecules, or physical systems. 
Let $\mathcal{G}$ be the set of all possible graphs $G$, which are defined by an unrestricted graph grammar \cite{rozenberg1997handbook}.  
The state space $\mathcal{S} \subset m\wp(\mathcal{G})$, where $m\wp$ denotes all sub-multisets formed by the elements of $\mathcal{G}$.\footnote{This extends power sets of sets to multisets.}

Actions, which transform states into new states, can then be seen as graph rewriting rules $p: L \leadsto R$, which means a graph $L$ is matched in the graphs in a state $s$ via (sub)graph isomorphism, cut out, and a different graph $R$ is glued in in this position.
\cite{rozenberg1997handbook,andersen2013generic}. This gives the learner a handle to induce the elements of the action space.

\subsection{World Programs}
Program induction is the task of generating a program consistent with a specification provided by input-output examples  \cite{gulwani2017program}.
A plethora of algorithms has been proposed for this task. Purely symbolic approaches are very data efficient, and can learn from very few examples, however, they tend to struggle in the face of noisy data. In contrast, approaches based on low-bias neural networks such as seq2seq-RNNs or transformers turn out to be quite robust to noise. However, without the adequate inductive biases they require large amounts of data to train, and do not generalise well beyond the training distribution. Also, several models combining neural and symbolic components have been proposed.\cite{gaunt2017differentiable, bovsnjak2017programming, evans2018learning,bunel2018leveraging}

In our formalism, we define the triple $(\mathcal{A}, \mathcal{A}(s), \mathcal{T})$ as the world program, which the learner needs to learn from observations $\mathcal{O}$ in the form of $(s_t, s_{t+1})$ pairs. 
We can then treat each graph-rewriting action $a_i \in \mathcal{A}$ as a subroutine, and $\mathcal{A}(s)$ as a map from a state to sets of subroutines $a_i$. Furthermore, we can leverage the induced actions $\mathcal{A}$ to learn the transition model $\mathcal{T}:\mathcal{S} \times \mathcal{A} \times \mathcal{S} \rightarrow [0,1]$ by sampling actions $a_i \thicksim \mathcal{A}$, and applying them to $\mathcal{O}$ to generate training examples consistent and inconsistent with the observations.

\begin{figure}[htbp]
\begin{center}
\includegraphics[width=\columnwidth]{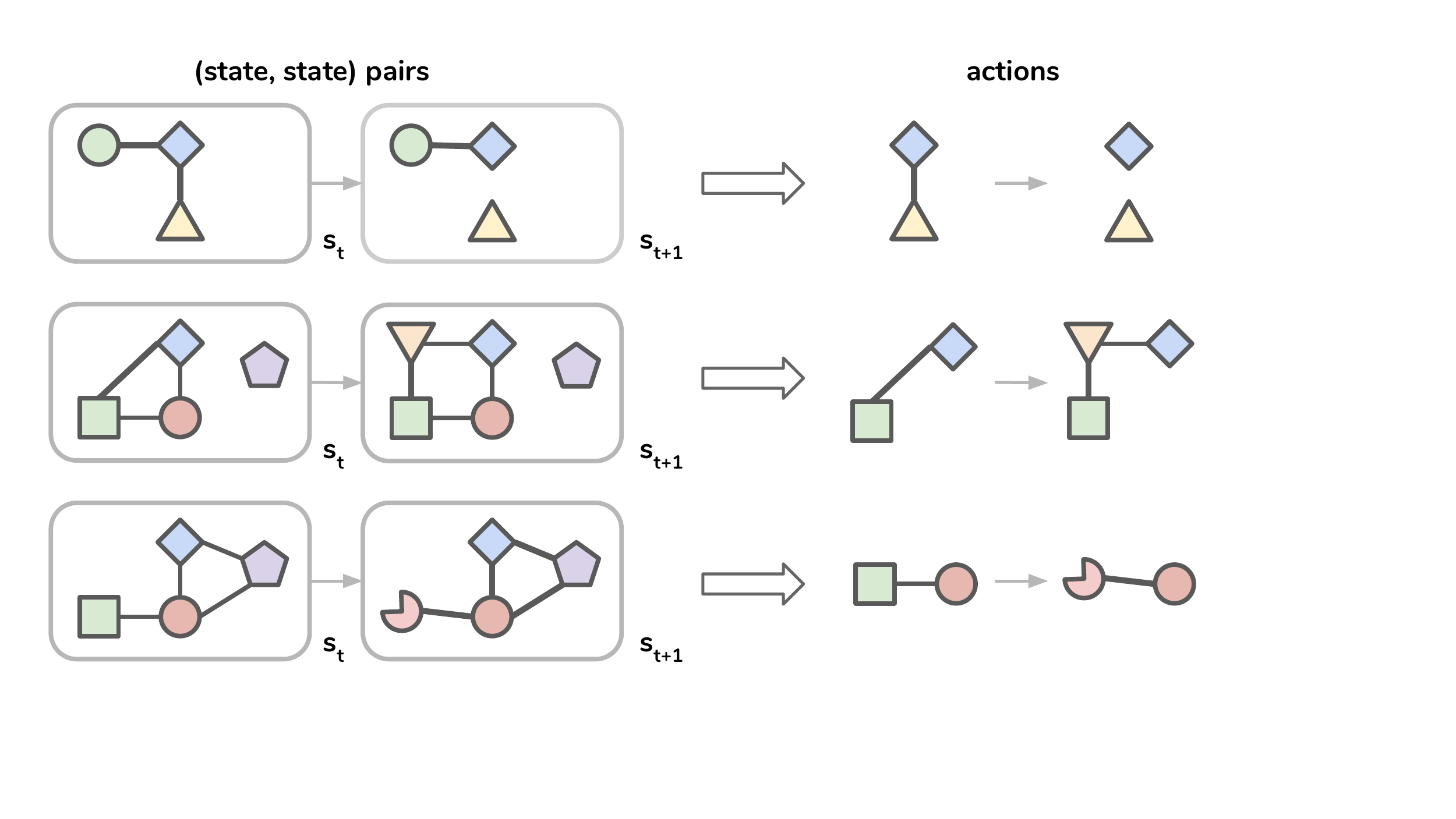}
\caption{Given subsequent state-state input-output pairs, the learner first has to induce the actions which gave rise to the transition, in the form of graph rewriting rules or subroutines of atomistic edit operations.}
\label{fig:indu}
\end{center}
\end{figure}

The concrete implementation of the program induction step is a hyperparameter. It can be realized by many different algorithms, using everything from purely symbolic to graph neural networks or purely distributed representations \cite{chemres,neusy,liu2017retrosynthetic, bradshaw2018generative, yin2018learning, bradshaw2019model}.

\subsection{Transition Functions and Policies}
Graph neural networks (GNNs) have been employed in chemistry since decades \cite{kireev1995chemnet,baskin1997neural,merkwirth2005automatic}, and can be traced back to algorithms for graph canonicalisation developed for chemical databases \cite{morgan1965generation,weisfeiler1968reduction} and path-based graph featurisation \cite{wiener1947structural}. GNNs have recently been shown to be useful as well in many other domains where relations between objects can be modelled as graphs, e.g. in physics and source code \cite{niepert2016learning,kipf2016semi, brockschmidt2018generative, bapst2019structured}. Here, to represent the states, graph neural networks provide the adequate inductive bias for the task.

\subsection{Planning with World Programs}
After the learner has induced the world program $(\mathcal{A}, \mathcal{A}(s), \mathcal{T})$, it can be employed in a simulator $\mathcal{M^\prime} = (\mathcal{S},\mathcal{A}, \mathcal{T},\mathcal{R})$ in combination with any reinforcement learning algorithm to train an agent or perform planning. The discrete nature of the problem and the availability of the model lends itself immediately to Monte Carlo Tree Search (MCTS) \cite{kocsis2006bandit,coulom2006efficient} and its extensions. 

\begin{figure}[htbp]
\begin{center}
\includegraphics[width=\columnwidth]{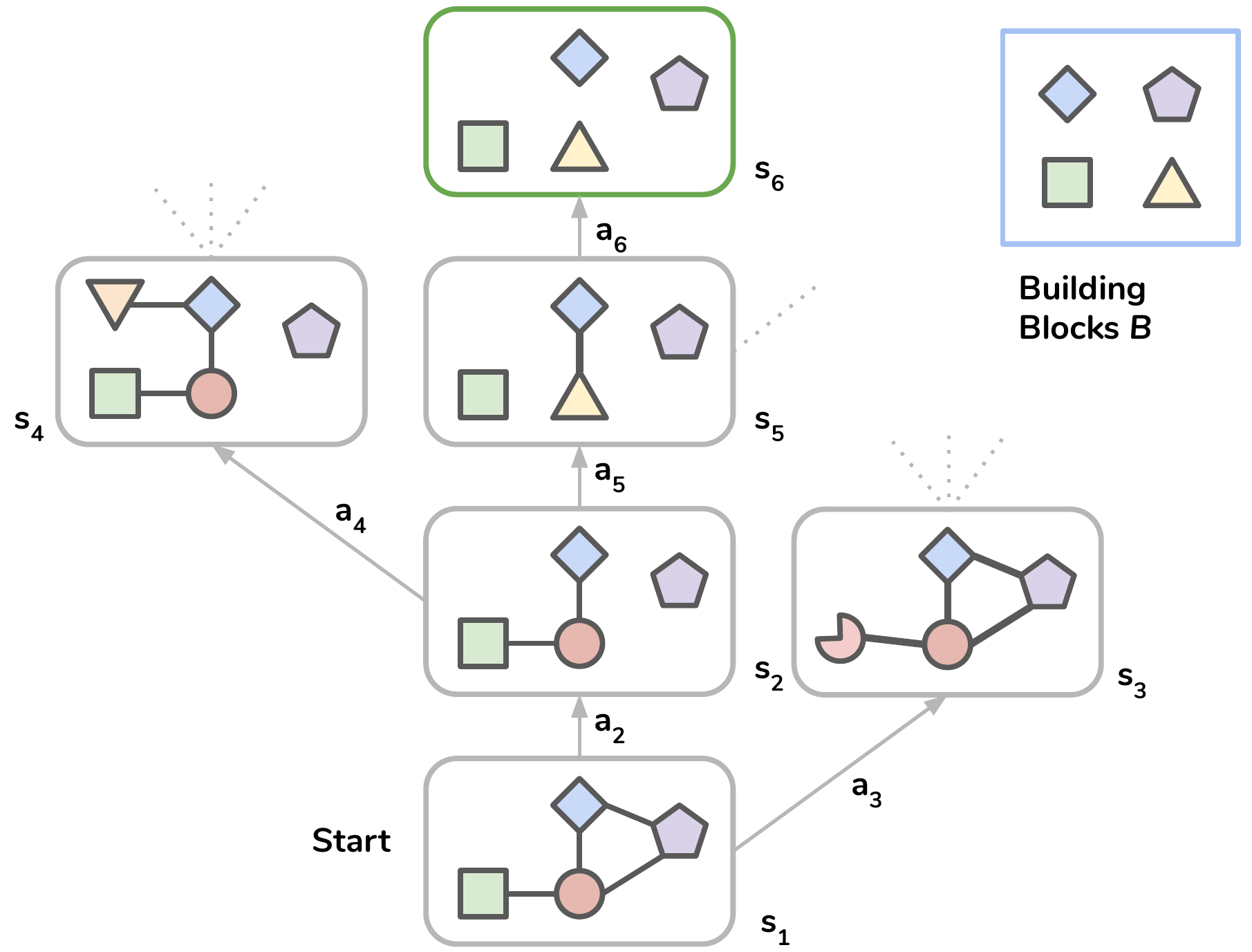}
\caption{States are multisets of graphs. The actions rewrite the graphs within a state by changing the vertices and edges. In this example, we consider the task of planning the construction of a complex object $s_1$ from a set of building blocks $\mathcal{B}$. State $s_6$ is a goal state as $\forall g_i \in s_6 : g_i \in \mathcal{B}$, and the branch from $s_1$ to $s_6$ is the solution/plan.}
\label{fig:bb}
\end{center}
\end{figure}

\section{Experiments}
The formalism described above was applied for the problem of chemical synthesis planning for small organic molecules, which are of central importance for human well-being as medicines or materials \cite{todd2005computer}. 
Here, starting with the target molecule we wish to make, the task is to perform a backward search by recursively deconstructing the molecule into increasingly simpler parts using formally reversed chemical reactions (graph rewriting rules), see Figure \ref{fig:bb}. The problem is solved iff the state is a subset of a predefined set of building blocks. More formally, given a set of building block graphs $\mathcal{B}$, a state $s_k$ is solved iff $\forall g_i \in s_k: m_i \cong b_j \in \mathcal{B}$, that is all graphs $m_i$ in $s_k$ are isomorphic to a building block graph $b_j \in \mathcal{B}$.
The results are reproduced from \cite{segler2018planning}, where this framework was first applied, however, formulated for a non-ML audience.

\subsection{Methods}
The graphs in this task are small organic molecules, consisting of atoms as vertices and bonds as edges. To encode these graphs, a special case of Graph Neural Networks (GNN) was used, where the trainable weights are replaced by hash functions, which drastically speeds up inference while performing well \cite{rogers2010extended}. 
To induce the world program, we first take chemical reactions reported in the literature as our ($s_t, s_{t+1}$) observations, with the reagents and the product of the reactions as the states. We then first employ a symbolic search procedure to delete all nodes and edges from the molecular graphs that do not get changed in the course of the reaction, to form $\mathcal{A}$, with $|\mathcal{A}| \approx 300,000$ actions.\footnote{This is 3 orders of magnitude larger than the game of Go.} To learn $\mathcal{A}(s)$, we train a neural network to predict the probability over all rules given s, and restrict the available actions to the top-k rules with a cumulative probability of $>99\%$.

The transition function $\mathcal{T}$ is a binary classifier which takes the same GNN as an input for states, and takes the difference of the embeddings of $s_t$ and $s_{t+1}$ as the representation of the action $a_t$. For more details, we refer the reader to \cite{segler2018planning}.

\subsection{Quantitative Results}
Table \ref{tab:exp} shows experimental results for several models in the task of predicting synthesis plans for 497 different molecules \cite{segler2018planning}. In PUCT-MCTS, the prior probability from the network representing $\mathcal{A}(s)$ is used to bias the exploration term. Monte Carlo Search (MCS) (sampling from the prior), UCT-MCTS, where the exploration term does not have a predicted probability contribution, and two Best First Search (BFS) variants all perform worse than PUCT-MCTS.\footnote{It has to be noted that we did not tune most of the hyperparameters (i.e. the world program induction algorithm, the neural network architectures of the policy network and the transition function) of PUCT-MCTS agent, except for the MCTS exploration constant. We therefore expect considerable gains by investigating these hyperparameters.}

\begin{table}[htb]
\caption{Experimental Results}
\begin{center}
\begin{tabular}{lrr}
\toprule
Agent & \% solved  & time (s/plan)  \\
\midrule
PUCT-MCTS  & 95.24 $\pm$0.09 &  13.0\\
MCS        & 89.54 $\pm$0.59 & 275.7\\
UCT-MCTS   & 87.12 $\pm$0.29 &  30.4\\
BFS (neural)       & 84.24 $\pm$0.09 &  39.1\\
BFS (heuristic) &    55.53 $\pm$2.02     &  422.1       \\
\bottomrule
\end{tabular}
\end{center}
\scriptsize{Time budget 3$\times$300~s and 100,000 iterations for MC(T)S or 3$\times$300~s and 100,000 expansions for BFS, per molecule ($3$ restarts were allowed). Std dev is given for the solved ratio. Results reproduced from \cite{segler2018planning}
}
\label{tab:exp}
\end{table}%

\subsection{Qualitative Results}
A notorious problem with model-based RL and planning is that strong learners can exploit imperfections in the simulator, leading to plans that do not translate to the real environment. 
Thus, to test the quality of the plans, we conducted two AB tests, with 45 expert organic chemists, who had to choose one of two plans in a double blind setup.

First, the subjects were presented with plans previously reported in the literature, and a plan generated by the PUCT-MCTS algorithm for the same target molecule.
The null hypothesis is that experts rate the computed plans as inferior (expected preference: computer 0\% vs literature 100\%). 
Surprisingly, we found that the experts perceived the computed routes (57.0\%) as equivalent to the literature routes (43.0\%). 

In the second test, the experts had to report their preferences for plans found by PUCT-MCTS or plans generated by a baseline system based on heuristic BFS. 
The subjects significantly preferred the routes generated by MCTS (68.2\%) over the baseline system (31.8\%).

\section{Related topics}
World program learning for discrete states and actions is related to the problem of inverse dynamics in continuous state and action spaces, e.g. for use in robotics, where the torques to produce a movement are induced from observed trajectories \cite{nguyen2010using,meier2016towards,christiano2016transfer}. How to extend the presented framework here to the continuous setting remains a topic for future work. 

Furthermore, there is a relation between world program induction and hierarchical reinforcement learning/option discovery \cite{RLbook}, which are predominantly used to extract spatiotemporal abstractions on multiple scales by joining sequences of (micro-)actions into more abstract macro actions, where both micro- and macro-actions having the same type or unit. 
In contrast, in a world program, the full action and its atomistic components do not have the same type. 

\section{Conclusion and Future Challenges}
Here, we have proposed a formalism to perform model-based planning in complex compositional environments, where the learner a priori does not have knowledge of the action space. As graph neural networks have progressed from chemical modeling problems to all kinds of relational prediction problems, we hope that the machine learning community might also take inspiration to build on techniques developed for chemical planning problems to other relational planning problems, such as automated theorem proving, code optimization/rewriting, or physical construction and deconstruction problems.

To conclude, we would also like to pose a small challenge: 
Given only a small set of historically played Go and Chess games as grid graphs, can you first learn the rules of the games to learn a simulator, and then compete with an AlphaZero-like agent \cite{silver2018general} using an agent trained in the learned simulator?

\section*{Acknowledgements}
The author thanks M. Ahmed, T. Edlich, and B. Fabian for feedback on the manuscript, and the anonymous reviewers for helpful suggestions.

\bibliography{wprg}
\bibliographystyle{icml2019}

\end{document}